\documentclass[10pt,twocolumn,letterpaper]{article}

\usepackage{cvpr}
\usepackage{times}
\usepackage{epsfig}
\usepackage{graphicx}
\usepackage{amsmath}
\usepackage{amssymb}
\usepackage{subfigure}
\usepackage{array}
\usepackage{multirow}
\usepackage{diagbox}
\usepackage{graphicx}
\usepackage{booktabs}
\usepackage{gensymb}
\usepackage{xspace}

\usepackage[breaklinks=true,bookmarks=false]{hyperref}
\usepackage{cleveref}

\newcommand{\OURS}{Neural Illumination\xspace}
\newcommand*{\Cdot}{\raisebox{-0.25ex}{\scalebox{1.75}{$\cdot$}}}
\newcommand{\mypara}{\vspace*{-3mm}\paragraph}

\cvprfinalcopy 

\def\cvprPaperID{1188} 

\ifcvprfinal\fi
\begin{document}

\title{
\OURS: Lighting Prediction for Indoor Environments
}

\author{Shuran Song \quad\quad  Thomas Funkhouser\\ Google and Princeton University}

\maketitle


\begin{abstract}
This paper addresses the task of estimating the light arriving from all directions to a 3D point observed at a selected pixel in an RGB image.   This task is challenging because it requires predicting a mapping from a partial scene observation by a camera to a complete illumination map for a selected position, which depends on the 3D location of the selection, the distribution of unobserved light sources, the occlusions caused by scene geometry, etc.  Previous methods attempt to learn this complex mapping directly using a single black-box neural network, which often fails to estimate high-frequency lighting details for scenes with complicated 3D geometry.  Instead, we propose ``\OURS,'' a new approach that decomposes illumination prediction into several simpler differentiable sub-tasks: 1) geometry estimation, 2) scene completion, and 3) LDR-to-HDR estimation.  The advantage of this approach is that the sub-tasks are relatively easy to learn and can be trained with direct supervision, while the whole pipeline is fully differentiable and can be fine-tuned with end-to-end supervision.   Experiments show that our approach performs significantly better quantitatively and qualitatively than prior work.  
\end{abstract}





\section{Introduction}

\begin{figure}[t]
    \includegraphics[width=\linewidth]{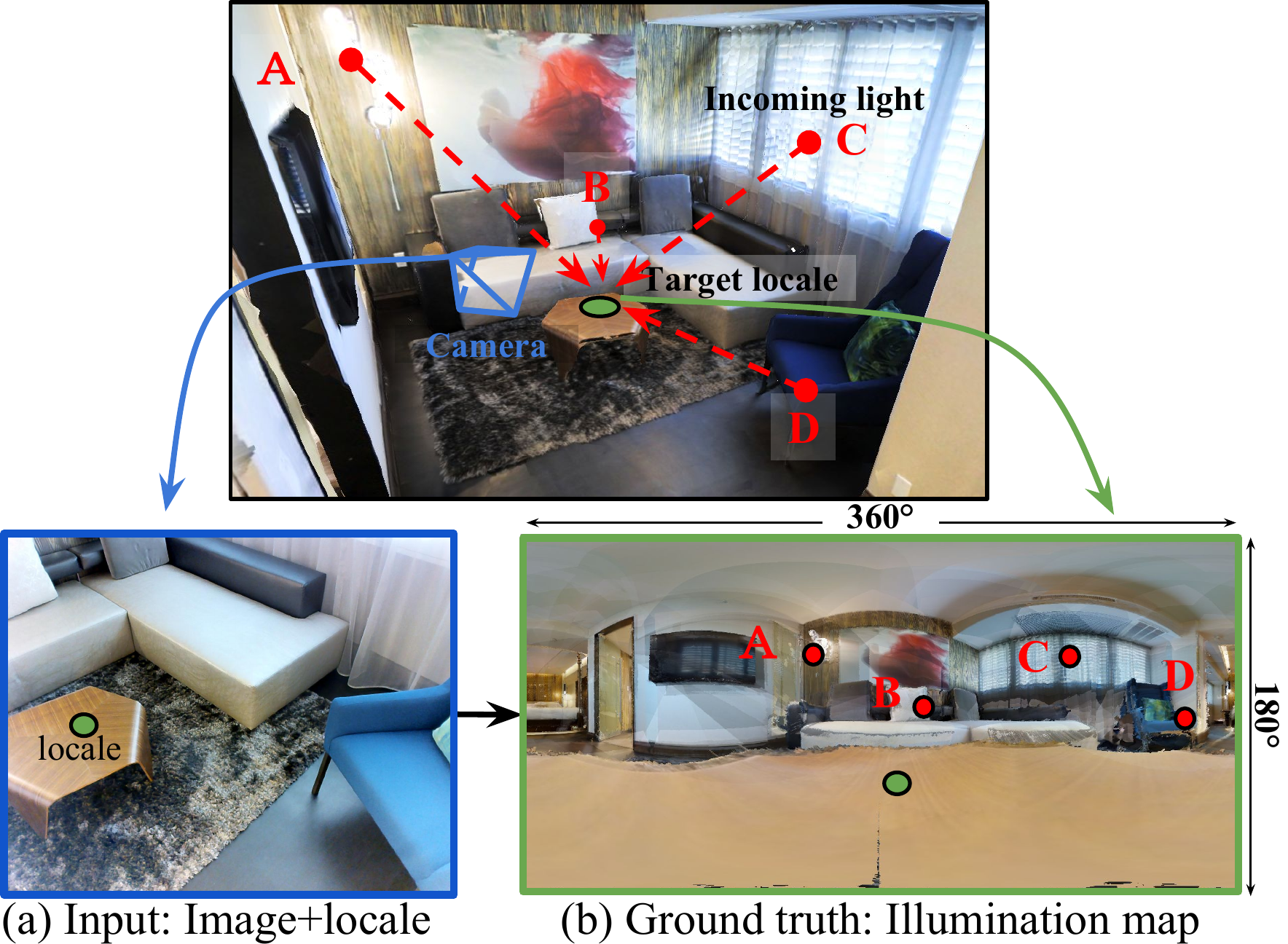}
    \vspace{1mm}
    \caption{Given a single LDR image and a selected 2D pixel, the goal of {\bf \OURS} is to infer a panoramic HDR illumination map representing the light arriving from all directions at the locale.   The illumination map is encoded as a spherical image parameterized horizontally by $\phi$ (0-360\degree) and vertically by $\theta$ (0-180\degree), where each pixel (\eg A,B,C,D) stores the RGB intensity of light arriving at the ``locale''  from the direction $(\phi,\theta)$. \label{fig:task}}
    \vspace{-3mm}
\end{figure}

The goal of this paper is to estimate the illumination arriving at a location in an indoor scene based on a selected pixel in a single RGB image.   As shown in Figure \ref{fig:task}(a), the input is a low dynamic range RGB image and a selected 2D pixel, and the output is a high dynamic range RGB illumination map encoding the incident radiance arriving from every direction at the 3D location (``locale'') associated with the selected pixel (Figure \ref{fig:task}(b)).  This task is important for a range of applications in mixed reality and scene understanding. For example, the output illumination map can be used to light virtual objects placed at the locale so that they blend seamlessly into the real world imagery (Figure \ref{fig:render}) and can assist estimating other scene properties, such as surface materials.

\begin{figure*}[t]
\vspace{-6mm}
    \includegraphics[width=\linewidth]{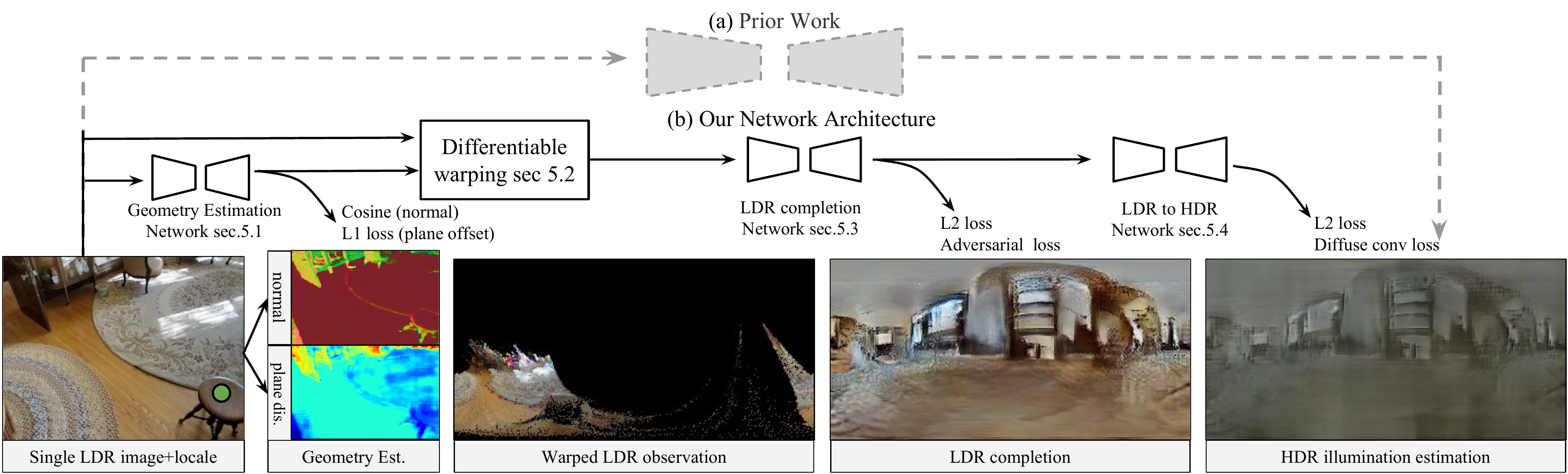}
    \caption{{\bf \OURS.}  In contrast to prior work (a) \cite{gardner2017learning} that directly trains a single network to learn the mapping from input images to output illumination maps, our network (b) decomposes the problem into three sub-modules: first the network  takes a single LDR RGB image  as input and estimate the 3D geometry of the observed scene. This geometry is used to warp pixels from the input image onto a spherical projection centered around an input locale. The warped image is then fed into LDR completion network to predict color information for the pixels in the unobserved regions. Finally, the completed image is passed through the LDR2HDR network to infer the HDR image. The entire network is differentiable and is trained with supervision end-to-end as well as for each intermediate sub-module.\label{fig:network}}
    \vspace{-3mm}
\end{figure*}

This goal is challenging because it requires a comprehensive understanding of the lighting environment. 
First, it requires understanding the 3D geometry of the scene in order to map between illumination observations at one 3D location (the camera) and another (the selected 3D locale).  Second, it requires predicting the illumination coming from everywhere in the scene, even though only part of the scene is observed in the input image (\eg the unobserved window in Figure \ref{fig:network}).  Third, it requires inferring HDR illumination from LDR observations so that virtual objects can be lit realistically.
While it is possible to train a single neural network that directly models the illumination function end-to-end (from an input LDR image to an output HDR illumination map) \cite{gardner2017learning}, in practice optimizing a model for this complex function is challenging, and thus previous attempts have not been able to model high-frequency lighting details for scenes with complicated 3D geometry.

In this paper, we propose to address these challenges by decomposing the problem into three sub-tasks. First, to estimate the 3D geometric relationship between pixels in the input image and the output illumination map, we train a network that estimates the 3D geometry from the observed image -- the estimated geometry is then used to warp pixels from the input image to a spherical projection centered at the target locale to produce a partial LDR illumination map.  Second, to estimate out-of-view and occluded lighting, we train a generative network that takes in the resulting partial illumination map and ``completes'' it -- i.e., estimates the LDR illumination for all unobserved regions of the illumination map.  
Finally, to recover high dynamic range information, we train another network that maps estimated LDR colors to HDR light intensities.  
All these sub-modules are differentiable.  They are first pre-trained individually with direct supervision and then fine-tuned end-to-end with the supervision of the final illumination estimation. 

Our \textbf{key idea} is that by decomposing the problem into sub-tasks, it becomes practical to train an end-to-end neural network -- each sub-module is able to focus on a relatively easier task and can be trained with direct supervision.
The first sub-task is of particular importance -- by predicting the 3D structure of the scene from the input image and using it to geometrically warp the input image such that it is spatially aligned with the output illumination map,
we are able to enforce pixel-to-pixel spatial correspondence between the input and output representations, which has previously been shown to be crucial for other dense prediction tasks, such as image segmentation and edge detection. 

To train and evaluate networks for this task, we have curated a benchmark dataset of paired input LDR images and output HDR illumination maps for a diverse set of locales in real-world scenes.  In contrast to prior work, our dataset leverages panoramas captured densely in real-world scenes with HDR color and depth cameras \cite{chang2017matterport3d}.  We use the depth channel to warp and resample those panoramas at arbitrary locales to produce a set of 90,280 ``ground truth'' illumination maps observed in 129,600 images.

The primary contribution of our paper is introducing an end-to-end neural network architecture for illumination estimation (\OURS) that decomposes the illumination estimation task into three sub-tasks. Our problem decomposition enables us 
1) to provide both direct intermediate and end-to-end supervision, and
2) to convert the input observation into an intermediate representation that shares the pixel-wise spatial correspondence with the output representation.
We show that this combination of neural network sub-modules leads to significantly better quantitative and qualitative results over prior work in experiments with our new benchmark dataset.


\section{Related Work}
Illumination estimation has been a long-standing problem in both computer vision and graphics. In this section, we briefly review work most relevant to this paper. 

\mypara{Capture-based Methods}
A direct way of obtaining the illumination of an environment is to capture the light intensity at a target location using a physical probe. 
Debevec \etal \cite{debevec1998rendering} first showed that photographs of a mirrored sphere with different exposures can be used to compute the illumination at the sphere's location. Subsequent works show that beyond mirrored spheres, it is also possible to capture illumination using hybrid spheres \cite{debevec2012single}, known 3D objects \cite{weber2018learning}, object's with know surface material \cite{Georgoulis_2017_ICCV}, or even human faces \cite{calian2018faces} as proxies for light probes.

However, the process of physically capturing high-quality illumination maps can be expensive and difficult to scale, especially when the goal is to obtain training data for a dense set of visible locations in a large variety of environments. 
In this paper, we propose to use existing large-scale datasets with RGB-D and HDR panoramas (Matterport3D \cite{chang2017matterport3d}) combined with image-based rendering methods to generate a large training set of high-resolution illumination maps in diverse lighting environments. 

\begin{figure}[t]

    \includegraphics[width=\linewidth]{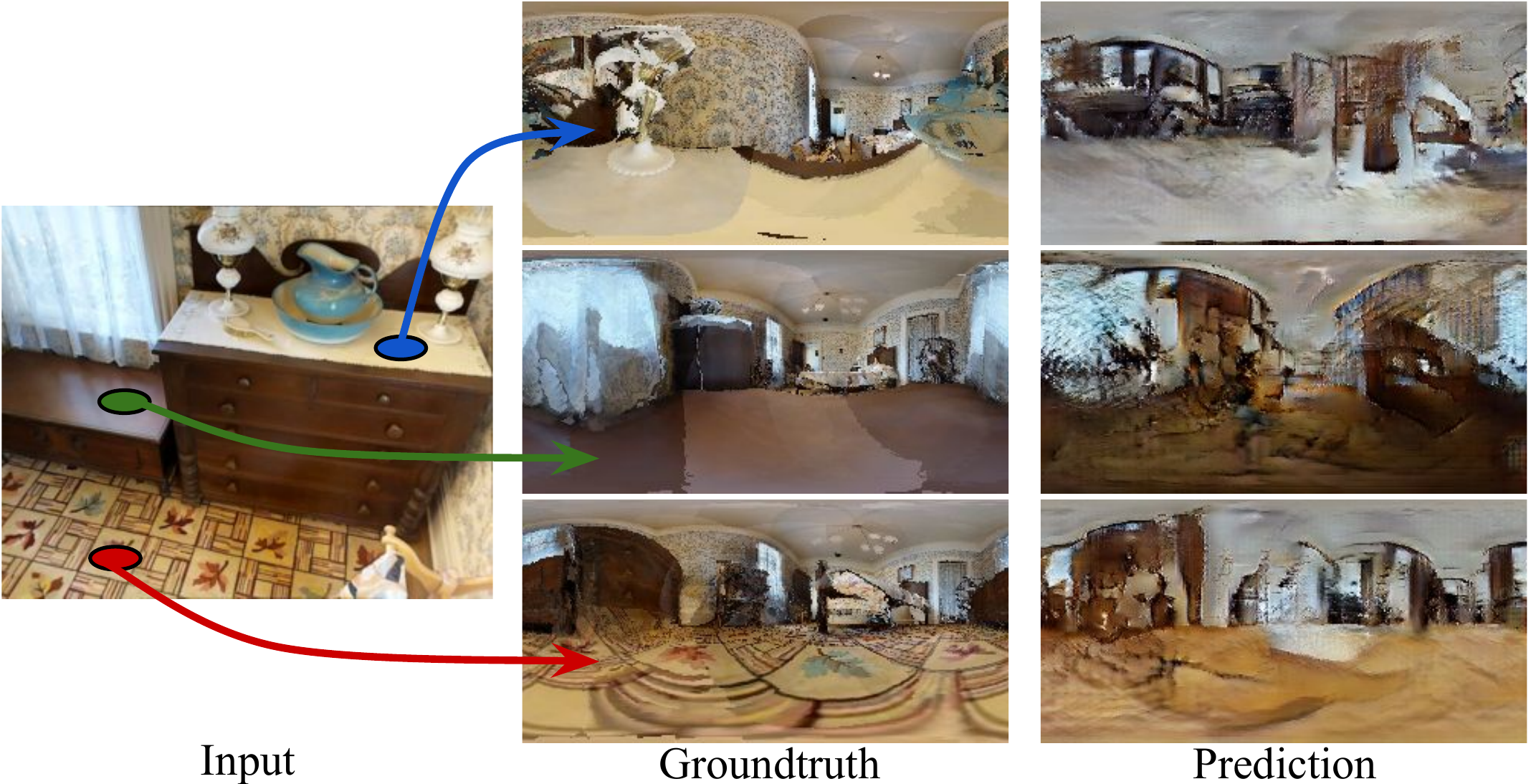}
    \caption{ \label{fig:diff_locale} \textbf{Spatially varying illumination.} By using the 3D geometry, we can generate ground truth illumination for any target locale. As a result, our model is also able to infer spatially varying illumination conditioned on the target pixel location. 
    }
    \vspace{-3mm}
\end{figure}

\mypara{Optimization-based Methods} 
One standard approach to estimating illumination is to jointly optimize the geometry, reflectance properties, and lighting models of the scene in order to find the set of values that best explain the observed input image.
However, directly optimizing all scene parameters is often a highly under-constrained problem -- an error in one parameter estimation can easily propagate into another. Therefore to ease the optimization process, many prior methods either assume additional user-provided ground truth information as input or make strong assumptions about the lighting models. For example, Karsch \etal \cite{karsch2011rendering} uses user annotations for initial lighting and geometry estimates. Zhang \etal \cite{zhang2016emptying} uses manually-annotated light-source locations and assumes knowledge of depth information. Lombardi and Nishino  \cite{lombardi2016reflectance} propose approximating illumination with a low-dimensional model, which subsequently has been shown to be sub-optimal for indoor scenes due to object reflective and geometric properties \cite{gardner2017learning}.   

There are also works that explore the idea that similar images share similar illumination estimates.
For example, Karsch \etal \cite{karsch2014automatic} uses image matching to find the most similar image crop from a panoramic database \cite{xiao2012recognizing} and then use the lighting annotations on those panoramic images to predict out-of-view light sources. Khan \etal \cite{khan2006image} directly flips observed HDR images to produce environment maps.
In contrast, our system does not require additional user inputs or manual annotations, and it does not make any explicit assumptions about the scene content or its lighting models. Instead, we enable our learning-based model to learn illumination priors directly from data.

\mypara{Learning-based Methods}
Deep learning has recently shown promising results on a number of computer vision tasks, including depth estimation \cite{liu2015deep,laina2016deeper} and intrinsic image decomposition \cite{zhou2015learning,li2018cgintrinsics}. Recently Gardner \etal \cite{gardner2017learning} propose to formulate the illumination estimation function as an end-to-end neural network. However, since the input and output representation of their network architecture does not share any immediate notion of pixel-to-pixel spatial correspondence, their model tends to generate illumination maps that reflect general color statistics of the training dataset, as opposed to important high-frequency lighting details.
In contrast, our model predicts the 3D geometric structure of the scene and uses it to warp the observed image into an intermediate representation that encodes the input information in a way that is spatially aligned to the output illumination map. This results in the ability to fully utilize input information and preserve high-frequency lighting details.
Moreover, Gardner \etal's algorithm does not generate illumination conditioned on a selected target pixel (i.e., it produces only one solution for each input image).  In contrast, our algorithm is able to recover the spatially varying illumination for any selected pixel (Figure \ref{fig:diff_locale}).

Apart from differences in network design, Gardner \etal also suffers from the lack of accurate ground truth training data.   Since their training data does not have depth, they use a sphere to approximate the scene geometry to warp a panorama to the target location.  Moreover, since most of their training data is LDR, they use a binary light mask to approximate the bright HDR illumination during pre-training.  While reasonable in the absence of 3D geometric and HDR information, these methods serve as weak approximations of ground truth.  We address both of these data issues by directly training our model on a dataset that has both accurate 3D geometry and illumination information for a dense set of observations. 

\begin{figure*}[t]
 \vspace{-5mm}
    \includegraphics[width=\linewidth]{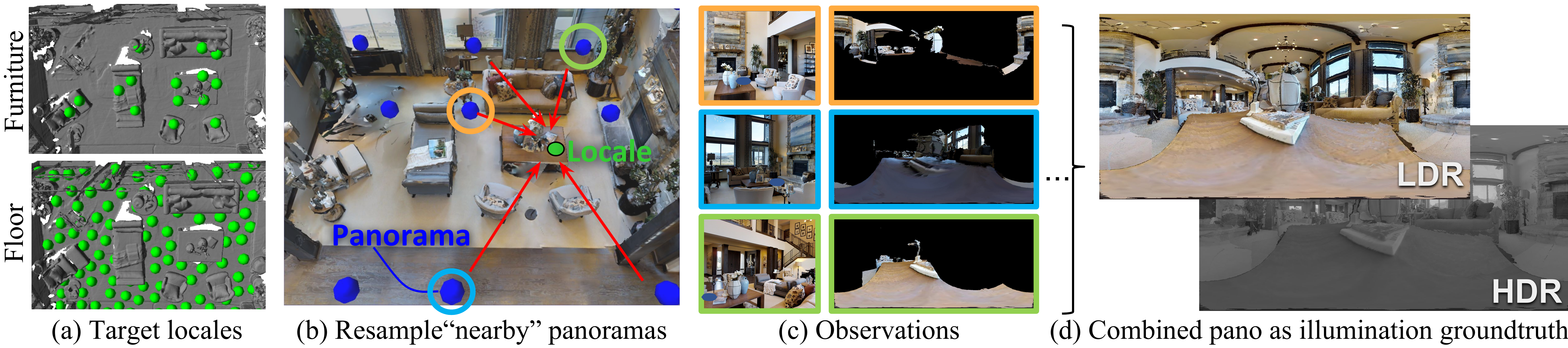}
    \caption{{\bf Ground truth illumination map generation.} We generate reconstructions of over 90 different building-scale indoor scenes using HDR RGB-D images from the Matterport3D dataset \cite{chang2017matterport3d}. From these reconstructions, we sample target locales (a) on supporting surfaces (floors and flat horizontal surfaces on furniture). For each locale, we use HDR and 3D geometric information from nearby RGB-D images to generate ground truth panoramic illumination maps.\label{fig:data}}
    \vspace{-3mm}
\end{figure*}

\section{Problem formulation}
We formulate illumination estimation as a pixel-wise regression problem modeled by a function $f$: $f(I|\ell)=H_{\ell}$ where $I$ is an input LDR image of a scene, $p$ is a selected pixel in the image.  $\ell$ is the 3D location of the pixel, and $H_\ell$ is the output HDR illumination around $\ell$. $H_\ell$ is represented as a spherical panoramic image with a 180\degree vertical FoV and 360\degree horizontal FoV. Each pixel $h(\phi,\theta) \in{H_\ell}$ of the panorama encodes the RGB intensity of incoming light to $\ell$  from the direction $(\phi,\theta)$.
We model $f$ as a feedforward convolutional neural network, the details of the network are described in Sec. \ref{sec:network}. We train $f$ on a large dataset of \{$I$,$\ell$\} and $H^*_\ell$ pairs generated from Matterport3D (Sec. \ref{sec:data}).

\section{Generating a Dataset of Illumination Maps}
\label{sec:data}
Obtaining a large dataset of ground truth illumination maps for training is challenging. On the one hand, using physical probes to directly capture illumination at a target locale \cite{debevec1998rendering,reinhard2010high,debevec2012single} provides accurate data, but scaling this capturing process across a diverse set of environments can be both costly and time-consuming. On the other hand, existing panoramic datasets (\eg \cite{xiao2012recognizing}) provide a simple way to obtain illumination maps, but only at the camera locations around which the panoramas were captured.

Instead, we propose to leverage the HDR RGB-D images from the Matterport3D dataset \cite{chang2017matterport3d} in combination with geometric warping to generate training data for arbitrary locales.  Matterport3D contains 194,400 registered HDR RGB-D images arranged in 10,800 panoramas within 90 different building-scale indoor scenes.
Since the panoramas provide ground truth HDR illumination maps for their center locations by direct observation, since they are acquired densely throughout each scene (separated by 2.5m or so), and since they have depth in addition to color, an RGB-D image-based rendering algorithm can reconstruct the illumination map {\em at any point in the scene} by warping and compositing nearby panoramas.   

The first step of our dataset curation process is to sample a set of target locales.   Ideally, the locales would cover the range of locations at which virtual objects could be placed in a scene.  Accordingly, we densely sample locations 10cm above the surface of the input mesh and create a new locale if a) it is supported by a horizontal surface ($n_z>$cos($\pi/8$)), b) the support surface has semantic label $\in \{$floor, furniture$\}$, c) there is sufficient volumetric clearance to fit an object with radius of 10cm, d) it is not within 50cm of any previously created locale.  For each locale, we backproject its location into every image $I$, check the depth channel to discard occlusions, and form a image-locale pair, \{$I$,$\ell$\}, for all others.

For each locale $\ell$, we construct an illumination map $H^*_{\ell}$ using RGB-D image-based rendering.  Though straightforward in principle, this process is complicated by missing depths at bright regions of the image (light sources, windows, strong specular highlights, etc.).  A simple forward projection algorithm based on observed depths would omit these important elements of the illumination map.   Therefore, we implemented a two-step process.   During the first step, we estimate the distance to the closest surface in every direction $d(\phi,\theta)$ by forward mapping the depth channel of every input image $I$ to $\ell$, remembering the minimum distance in every direction, and filling holes with interpolation where no samples were mapped. Then, we reconstruct the illumination map for $\ell$ by resampling the HDR color channels of the input images via reverse mapping and blending the samples with weights proportional to $1/d^4$, where $d$ is the distance between the camera and the locale.  This process produces illumination maps with smooth blends of pixels from the nearest panoramas with holes filled by other panoramas further away.
Overall, we generate 90,280  locales and 360,432
\{$I$,$\ell$\} and $H^*_{\ell}$  pairs using this process Figure \ref{fig:data} (a) shows examples for one scene. 

\begin{figure}[t]
    \includegraphics[width=\linewidth]{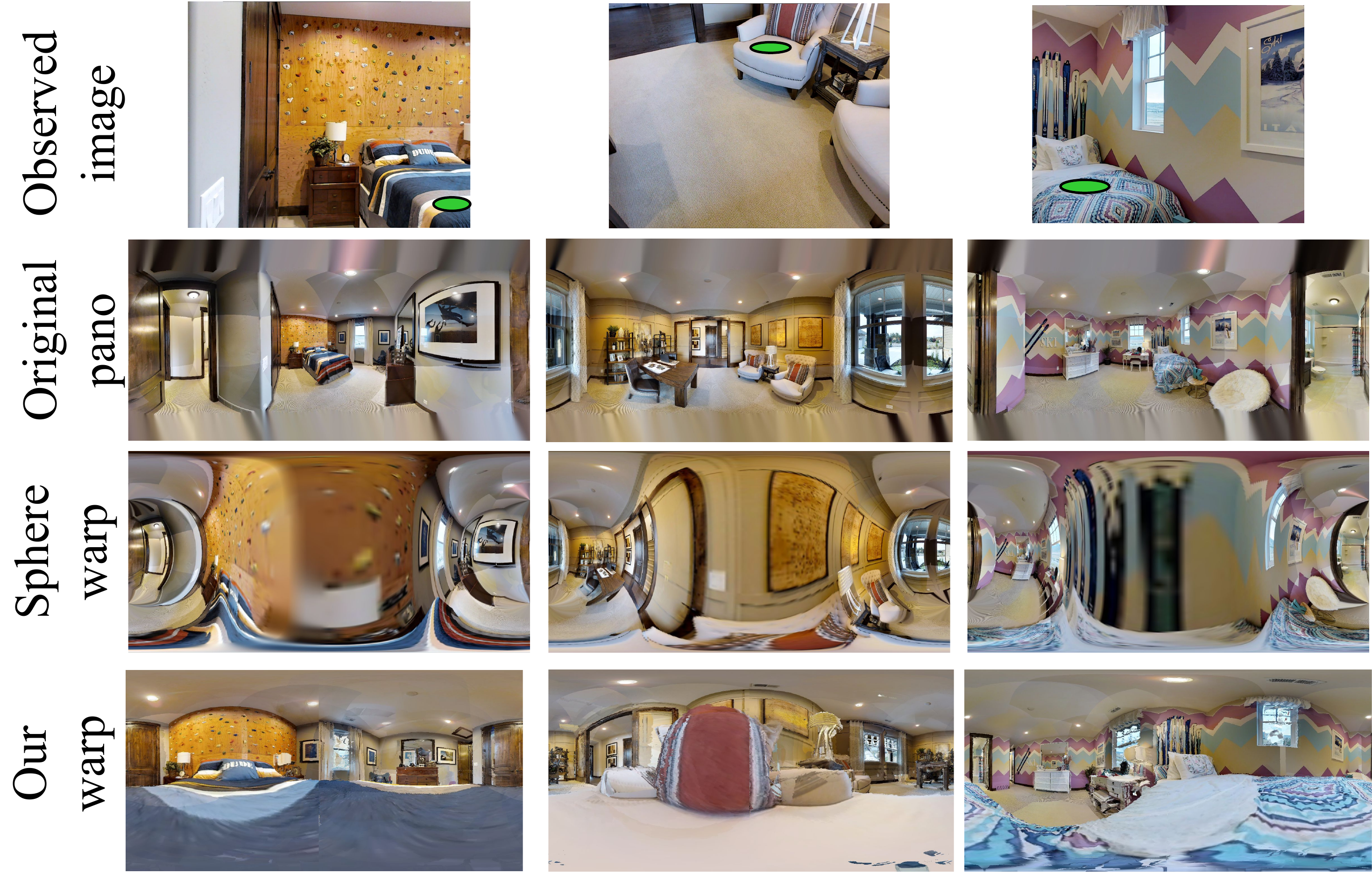}
    \caption{\label{fig:warping} {\bf Comparison of warping methods.} In our data generation process, we use 3D scene geometry to generate geometrically accurate ground truth illumination, which accounts for complex geometric structures and therefore more accurate than using spherical warping from 2D panoramas as in \cite{gardner2017learning}. }
    \vspace{-3mm}
\end{figure}

Though the illumination maps produced this way are not always perfect (especially for highly specular surfaces), they have several favorable properties for training on our task. 
First, they are sampled from data collected by a large number of photographers \cite{chang2017matterport3d} (mostly for real estate applications), and thus they contain a diverse set of lighting environments that would be difficult to gain access to otherwise.
Second, they provide a unique illumination map for each 3D locale in a scene.  Since multiple locales are usually visible in every single image, the dataset supports learning of spatial dependencies between pixel selections and illumination maps.  For example, Figure \ref{fig:diff_locale} shows that our network is able to infer different illumination maps for different pixels selections in the same input image. 
Third, the ``ground truth'' illumination maps produced with our RGB-D warping procedure are more geometrically accurate than others produced with spherical warping \cite{gardner2017learning}.  As shown in Figure \ref{fig:warping}, our warping procedure is able to account for complex geometric structures and occlusions in the scene.


\section{Network Architecture}
\label{sec:network}
In this section, we describe the convolutional neural network architecture used to model $f$, which consists of four sequential modules: 1) a geometry (RGB-to-3D) estimation module, 2) a differential warping module which warps the input RGB observation to the target locale using the estimated 3D information, 3) an out-of-view illumination estimation module, and 4) an LDR-to-HDR estimation module. Each module is pre-trained individually with its input and output pairs derived from ground truth information. Then all the modules are fine-tuned together end-to-end. Figure \ref{fig:network}, shows the network architecture.
By decomposing the network into sub-modules we allow each sub-module to focus on a relatively easier task with direct supervision. We find that providing both intermediate and end-to-end supervision is crucial for efficient learning.

\subsection{Geometry Estimation}
The geometry estimation module takes a single RGB image $I$ as input and outputs a dense pixel-wise prediction of the visible 3D geometry $G_I$.
Similar to Song \etal \cite{song2016im2pano3d}, $G_I$ is represented with a ``plane equation'' for each pixel. Specifically, we feed $I$ through a two-stream fully convolutional U-Net \cite{ronneberger2015u} to infer pixel-wise predictions of surface normals and plane offsets (\ie distance-to-origin). We then pass both predicted outputs through a differentiable PN-layer \cite{song2016im2pano3d} to convert the estimated surface normals and plane distances into a pixel-wise prediction of 3D locations. Direct supervision is provided to the 1) surface normal predictions via a cosine loss, 2) plane offset predictions via an $\ell1$ loss, and 3) final 3D point locations via an $\ell1$ to ensure consistency between the surface normal and plane offset predictions. Training labels are automatically obtained from the 3D data available in the Matterport3D dataset \cite{chang2017matterport3d}.
As shown in \cite{song2016im2pano3d}, this output representation provides strong regularization for large planar surface and is therefore able to produce higher quality predictions than directly predicting raw depth values \cite{eigen2014depth,laina2016deeper}.
At the same time, it also maintains the flexibility of representing any surfaces -- i.e., is not limited to a fixed number of planar surfaces, as in \cite{liu2018geometry}).

\subsection{Geometry-aware Warping}
The next module uses the estimated scene geometry $G_I$ to map the pixels in the input image $I$ to a panoramic image $\phi_\ell$ representing the unit sphere of rays arriving at $\ell$. We do this warping through a forward projection using the estimated scene geometry and camera pose.
The unit sphere projection that defines the panorama $\phi_\ell$ is oriented upright along $n_\ell$, which should be aligned with the gravity direction assuming that $\ell$ lays on a supporting surface (\eg floors and flat horizontal surfaces on furniture). 
Image regions in $\phi_\ell$ that do not have a projected pixel are set to -1. 
The resulting warped input observation is a panorama image with missing values that shares a pixel-wise spatial correspondence to the output illumination map. 
Since this warping module is entirely differentiable, we implement it as a single network layer.

\subsection{LDR Panorama Completion}
The third module takes the mapped observed pixels of $\phi_\ell$ as input and outputs a dense pixel-wise prediction of illumination for the full panoramic image $\psi_\ell$ including both observed and unobserved pixels. 
$\psi_\ell$ is represented as a 3-channel LDR color panorama. 

One of the biggest challenges for out-of-view illumination estimation comes from the multi-modal nature of the problem -- there can be multiple possible solutions of $\psi_\ell$ with illumination patterns that result in similar observations. 
Therefore, in addition to providing only pixel-wise supervision, we train this module with adversarial loss using a discriminator network \cite{goodfellow2014generative,isola2017image}.
This adversarial loss provides a learnable high-level objective by learning a loss that tries to classify if the output image is real or fake, while simultaneously training the generative model to minimize this loss. 
Our experiments show that this adversarial loss enables the network to produce and more realistic illumination outputs with sharper and richer details. 

This module is implemented as a fully convolutional ResNet50 \cite{he2016deep}. Since both the input and output of this module are represented as spherical panoramic images, we utilize distortion-aware convolutional filters that account for the different spherical distortion distributions for different regions of the image \cite{tateno2018distortion}.
This distortion-aware convolution resamples the feature space according to the image distortion model in order to improve the translational invariance of the learned filters in the network. 

\begin{figure}[t]
    \includegraphics[width=\linewidth]{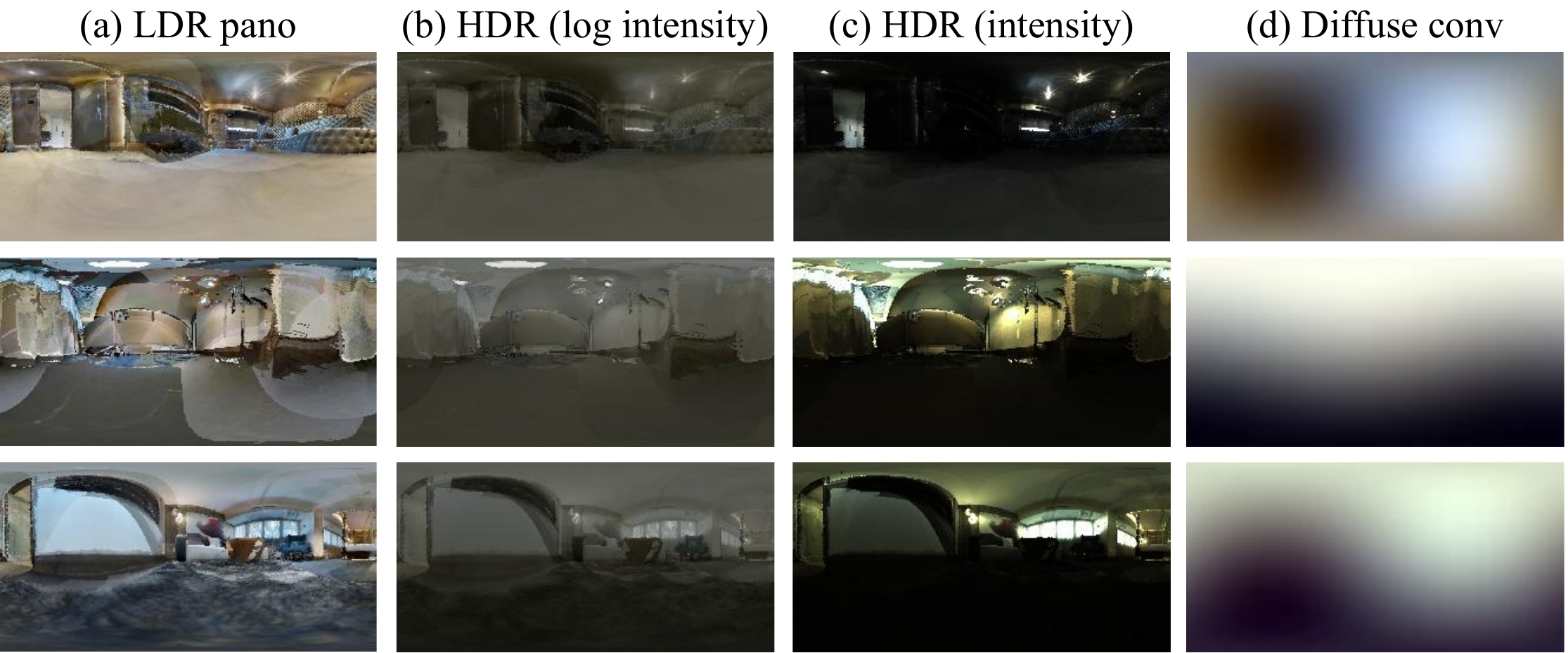}
    \caption{\label{fig:diffconv} Examples of a) LDR image, b) log scaled HDR $J$, c) HDR intensity $H$, d) diffuse convolution of HDR intensity $D(H)$.}
    \vspace{-3mm}
\end{figure}

\subsection{LDR-to-HDR Estimation}
The final module takes the predicted LDR illumination as input and outputs a dense pixel-wise prediction of HDR illumination intensities representing incident radiance in every direction at $\ell$.  This prediction is important because LDR images may have intensity clipping and/or tone-mapping, which would not be suitable for lighting virtual objects. 

Like Eilertsen \etal \cite{eilertsen2017hdr}, we formulate the LDR-to-HDR estimation as a pixel-wise regression problem, but instead of predicting the HDR value for only bright pixels and using a fixed function to map the rest of the pixels, our LDR-to-HDR module learns the mapping function for all pixels from the LDR space to the HDR space. 
The module is trained with supervision from: 1) a pixel-wise $\ell2$ loss $L_{\ell2}$,  and 2) a diffuse convolutional loss $L_d$. 

The pixel-wise $\ell2$ loss measures the visual error when re-lighting a perfectly \emph{specular} surface at $\ell$:

{\footnotesize
\vspace{-2mm}
$$L_{\ell2} =\frac{1}{N}\sum_{i=1}^{N}(J(i)-J^*(i))$$
}

\noindent 
where the $J$ is log-scaled image of the final light intensity $H$, defined as:

{\footnotesize
\[
    H(i)= 
\begin{cases}
    J(i) * 65536*8e^{-8} ,& J(i) \leq 3000\\
     2.4e^{-4} * 1.0002^{(J(i)*65536 - 3000)} ,& J(i) > 3000\\
\end{cases}
\]
}

The diffuse convolutional loss measures the visual error when re-lighting a perfectly \emph{diffuse} surface:
{\footnotesize 
$$L_d =\frac{1}{N}\sum_{i=1}^{N}(D(H(i))-D(H^*(i)))$$
} \vspace{-1mm}
where $D$ is the diffuse convolution function defined as:
{\footnotesize
$$ D(H,i) = \frac{1}{K_i} \sum_{\omega \in \Omega_i } H(\omega)s(\omega)(\omega\cdot\vec{n_i}) $$
}
and $\Omega_i$ is the hemisphere centered at pixel $i$ on the illumination map, $\vec{n_i}$ the unit normal at pixel $i$, and $K_i$ the sum of solid angles on $\Omega_i$. $\omega$ is a unit vector of direction on $\omega_i$ and $s(\omega)$ the solid angle for the pixel in the direction $\omega$.
This loss function is similar to the ``cosine loss'' function proposed by Gardner \etal \cite{gardner2017learning}, but rather than progressively increasing the Phong exponent value during training, we keep the Phong exponent value equal to 1.
In our implementation, we reduce the memory usage by computing $L_d$ on a downsized illumination map with average pooling.  

The final loss is computed as $L = \lambda_1L_{\ell2} + \lambda_2L_d $, where $\lambda_1 = 0.1$ and  $\lambda_2 = 0.05$. 
By combining these two losses during training, we encourage our model to reproduce both low and high frequency illumination signals. Figure \ref{fig:diffconv} shows examples of HDR images and their diffuse convolution.

\section{Evaluation}
We train and test our algorithm on the data generated from Section \ref{sec:data}, using the train/test split provided by the Matterport3D dataset \cite{chang2017matterport3d}.   The following experiments investigate qualitative and quantitative comparisons to prior work and results of ablation studies.  More results and visualizations can be found in the supplementary material.


\mypara{Evaluation metrics.}
We use the following evaluation metrics to quantitatively evaluate our predicted illumination maps $H_\ell$:

\begin{itemize}
\setlength{\topsep}{0pt}
\setlength{\parsep}{0pt}
\setlength{\parskip}{0pt}
\setlength{\itemsep}{2pt}

\item[$\Cdot$] {\bf Pixel-wise $\ell2$ distance error} is the sum of all pixel-wise $\ell2$ distances between the predicted $H_\ell$ and ground truth $H^*_\ell$ illumination maps. $\ell2 (\log)$ computes the $\ell2$ distance in the $\log$ intensity. Intuitively, this error measures the approximate visual differences observed when the maps are used to render a perfectly \textit{specular} surface at the target locale.

\item[$\Cdot$] {\bf Pixel-wise diffuse convolution error} is the sum of all pixel-wise $\ell2$ distances between $D(H_\ell)$ and $D(H^*_\ell)$. This error measures the approximate visual differences observed when the maps are used to render a perfectly \textit{diffuse} surface at the target locale.
\end{itemize}

\begin{figure*}[t]
\vspace{-3mm}
    \includegraphics[width=\linewidth]{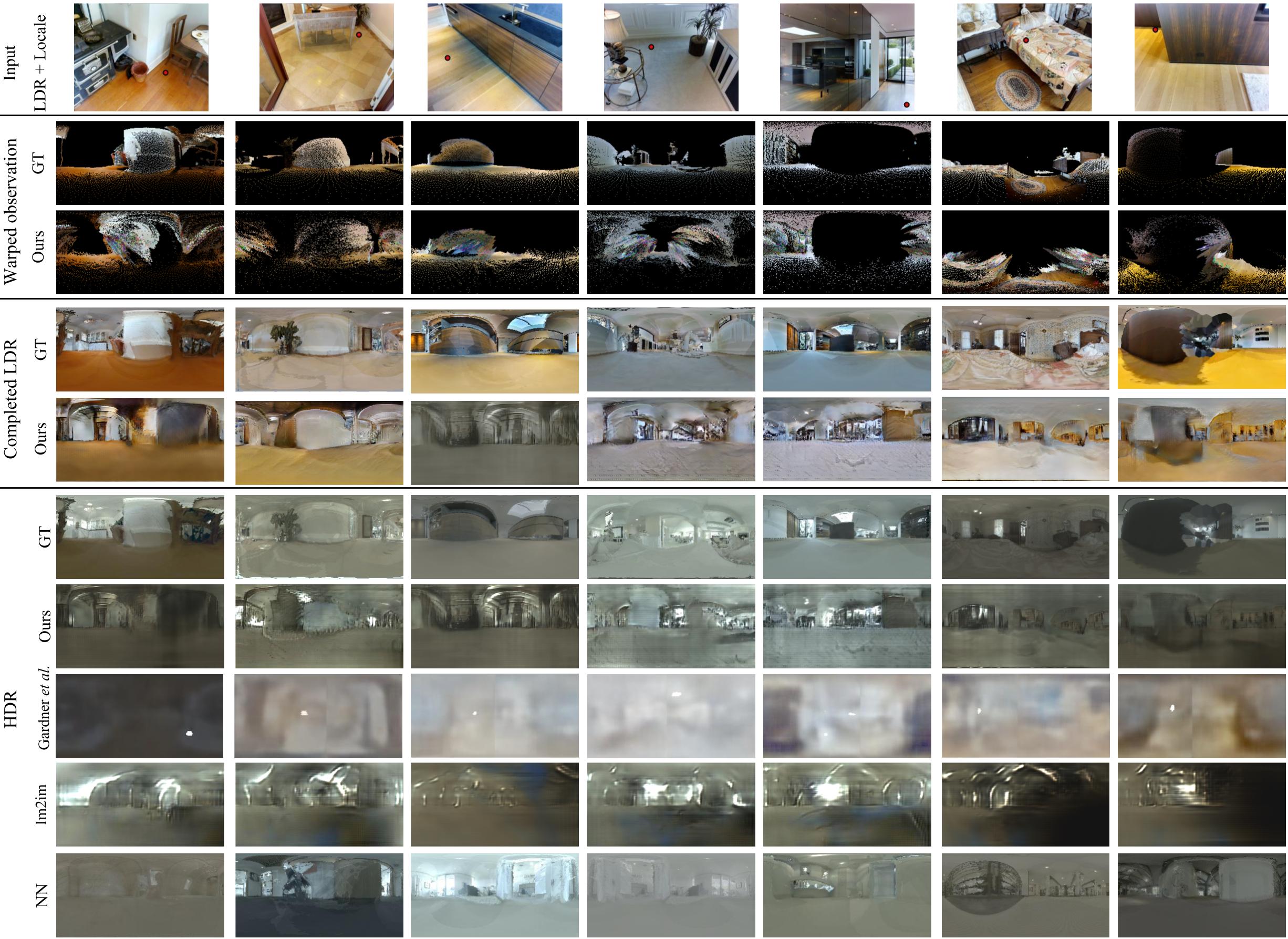}
    \caption{{\bf Qualitative Results} \label{fig:result} 
    (Row 1) show the input image and selected locale. (Row 2,3) show the warped observation using ground truth an predicted geometry. (Row 4,5) show the completed LDR. (Row 6-10) show the final HDR illumination visualized with gamma correction ($\gamma$=3.3). 
    We can  observe that the illumination maps estimated by our approach are more accurate and also contain richer high frequency details.
    }
    \vspace{-3mm}
\end{figure*}

\begin{figure*}[t]
    \includegraphics[width=\linewidth]{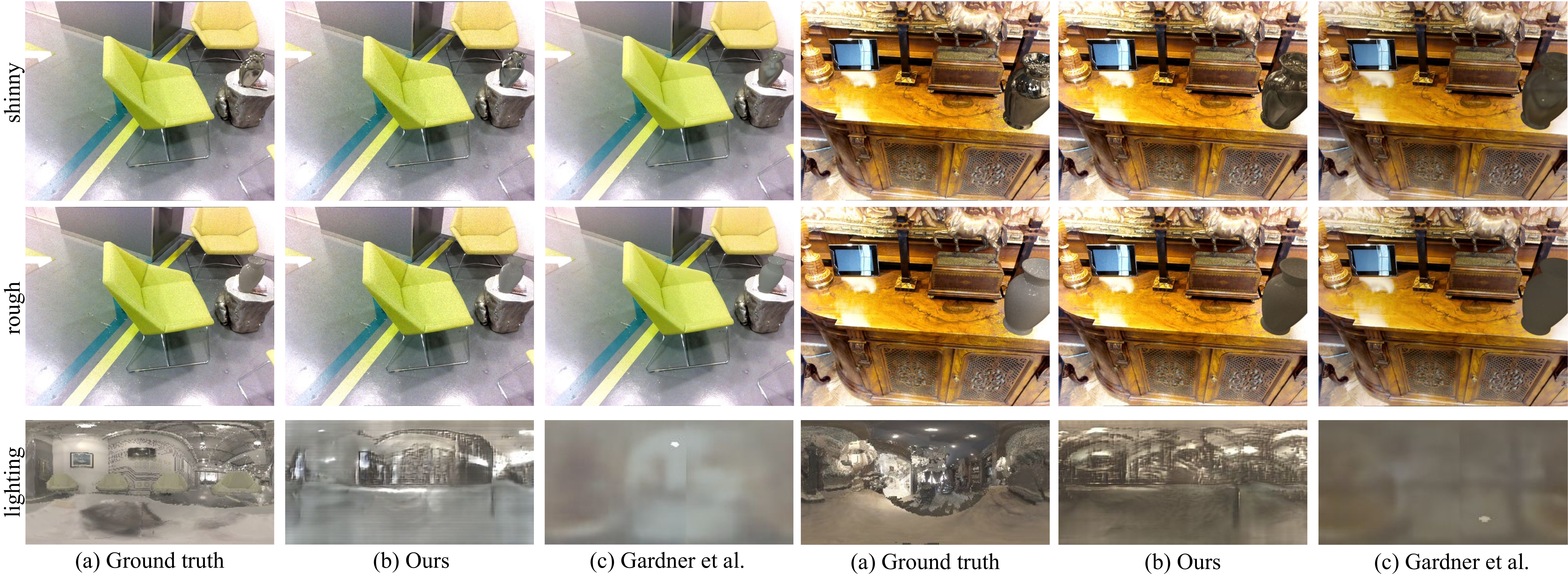}
    \caption{\label{fig:render} {\bf Object relighting example.} Here we show qualitative comparisons of relighting results rendered by Mitsuba using the illumination maps from (a) ground truth, (b) our algorithm, and (c) Gardner \etal.  We show images rendered with two different surface materials composited over the original observations (the first and second rows) and the illumination maps (third row). 
    Compared to (c), our algorithm is able to produce output illumination maps that contain much richer and more realistic high frequency detail.
    Of course, it also makes mistakes, for example by predicting extra light sources on the ceiling, which is incorrect but still plausible given the observation.
    }
    \vspace{-3mm}
\end{figure*}

\mypara{Comparisons to state-of-the-art.}
Table \ref{table:compare} shows quantitative comparisons of our approach to two alternative baselines: 1) Gardner \etal \cite{gardner2017learning}, and 2) a nearest neighbour retrieval method.  
Gardner \etal estimates the illumination condition of a given input image by training a single convolutional neural network with end-to-end supervision. We rotated each of the predicted panorama along the x-axis to align with ground truth coordinate frame before evaluation. 
Row 1 of Table \ref{table:compare} shows the performance of Gardner \etal's model trained on their original LRD+HDR panorama dataset and tested on our test set. 
We also re-implement an image-to-image prediction network that is similar to Gardner \etal's model and train it directly on our training data (LDR and full HDR illumination pairs) to remove potential dataset biases.
This model [Im2Im network] achieves better performance than the original model but is still less accurate than ours.
With a qualitative comparison (Figure \ref{fig:result},\ref{fig:render}), we can observe that by estimating and utilizing the 3D scene geometry, our algorithm is able to produce output illumination maps that contain much richer and more realistic high frequency details. 
Moreover, Gardner \etal's algorithm does not allow users to input a specific target pixel -- i.e., they generate only one lighting solution for each input image. In contrast, our algorithm is able to recover the spatially varying lighting distribution for any selected locale in the image, which can be quite different from one another (Figure \ref{fig:diff_locale}). 

\begin{table}[h]
\centering
 \vspace{-2mm}
\begin{tabular}{l|c|c|c}
\hline
Method   & $\ell2 (\log)$ &  $\ell2$  & diffuse \\ \hline
Gardner \etal \cite{gardner2017learning}          &0.375 &0.977 & 1.706\\
Im2Im network &0.229 & 0.369& 0.927\\
Nearest Neighbour                                 &0.296 & 0.647& 1.679 \\ 
Ours                                              &\textbf{0.202} & \textbf{0.280} & \textbf{0.772}\\ \hline
\end{tabular}
\vspace{1mm}
\caption{Comparing the quantitative performance of our method to that of Gardner \etal \cite{gardner2017learning} and a nearest neighbour retrieval method.  \label{table:compare}}
\vspace{-5mm}
\end{table}

\mypara{Modularization v.s. additional supervision.}
While we show that our network is able to achieve better performance than the single end-to-end model, it is still unclear whether the performance gain comes from the additional supervision or the network modularization. 
To investigate this question, we trained an end-to-end network that takes in a single LDR image as input and directly outputs the completed 3D geometry, LDR images, and HDR images at the final layers. This network is trained with supervision for all three predictions but does not have any network decomposition.
Table \ref{table:decom} shows the results. 
The performance gap between this network and ours demonstrates that naively adding all of the available supervision at the end of the network without proper network modularization and intermediate supervision does not work as well as our approach and generates significantly lower-quality illumination maps.

\begin{table}[h]
\vspace{-2mm}
\centering
\begin{tabular}{l|c|c|c}
\hline
 & $\ell2 (\log)$ & $\ell2$  & diffuse \\ \hline
without  &0.213 &0.319 &0.856\\ 
with (ours) &\textbf{0.202} & \textbf{0.280} & \textbf{0.772}\\ \hline
\end{tabular}
\vspace{1mm}
\caption{\label{table:decom} Effects of modularization.}  
\vspace{-3mm}
\end{table}

\mypara{Comparisons to variants with oracles.}

To study how errors in intermediate predictions impacts our results, we execute a series of experiments where some data is provided by oracles rather than our predictions.  In the first experiment, we trained a network that takes as input a LDR image already warped by a depth oracle and omits the first two modules (LDR+D$\rightarrow$HDR).  In a second experiment, we trained a version of that network that instead inputs a warped HDR image and omits execution of the last module.  These networks utilize ground truth data, and thus are not fair comparisons.  Yet, they provide valuable information about how well our network possibly could perform and which modules contribute most to the error.  Looking at Table \ref{table:input}, we see that providing ground truth depth improves our algorithm marginally (e.g., $\Delta\ell2=0.011$), while also providing ground truth HDR improves it more (e.g., $\Delta\ell2=0.043$).  We conjecture it is because errors are concentrated on bright light sources.  Overall, the performance of our algorithm is about halfway between the best oracled version [HDR+D$\rightarrow$HDR] and the baselines in Table \ref{table:compare}.

\begin{table}[h]
\vspace{-1mm}
\centering
\begin{tabular}{l|c|c|c}
\hline
Method   & $\ell2 (\log)$ &  $\ell2$  & diffuse \\ \hline
LDR$\rightarrow$HDR   &0.202 & 0.280 & 0.772\\ 
LDR+D$\rightarrow$HDR &0.188 & 0.269 & 0.761\\  
HDR+D$\rightarrow$HDR &\textbf{0.131} & \textbf{0.212} & \textbf{0.619}\\ \hline
\end{tabular}
\vspace{1mm}
\caption{Comparisons to variants with oracles. \label{table:input} }
\vspace{-3mm}
\end{table}

\mypara{Effects of different losses.}
To study the effects of different loss functions, we evaluate 
the performance of the model ``HDR+D$\rightarrow$HDR'' using different combinations of loss functions.
Figure \ref{fig:losses} shows quantitative and qualitative results.
From the results we can observe that with only an $\ell2$ loss, the network tends to produce very blurry estimations that are close to the mean intensity of the input images. By adding the adversarial loss, the network starts to be able to infer more realistic high frequency signals and spotlights, but also introduces additional noises and errors in the prediction.
By further adding a diffuse convolution loss [l2+gan+df], the network is able to predict overall more accurate illumination especially for the high intensity areas. 

\begin{table}[h]
\centering
\includegraphics[width=\linewidth]{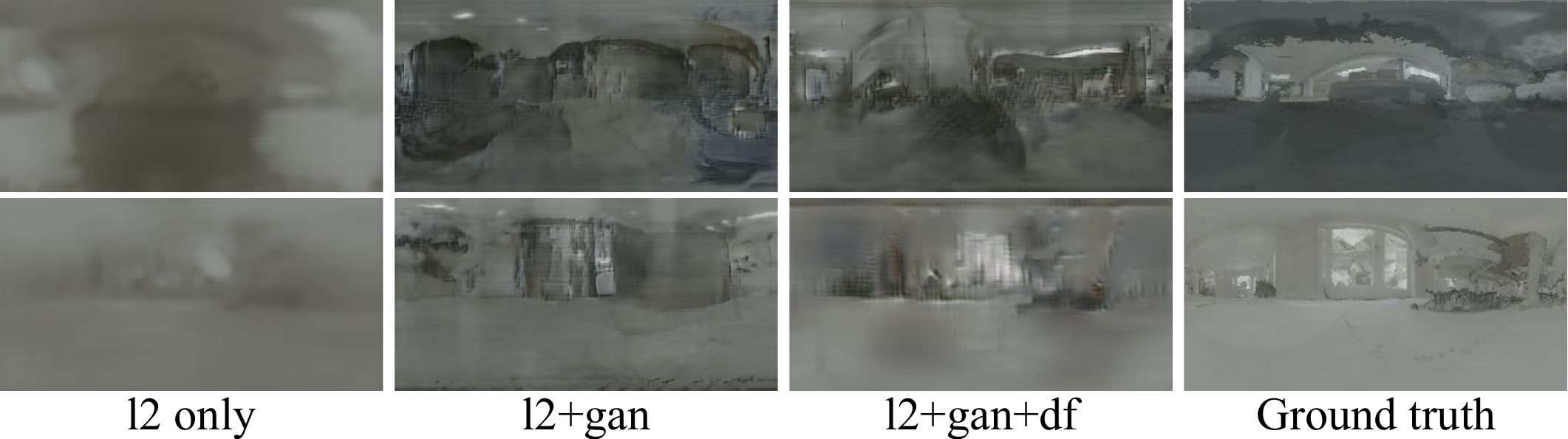}\\
\vspace{2mm}
\begin{tabular}{l|c|c|c}
\hline
loss   & $\ell2 (\log)$ & $\ell2$  & diffuse \\ \hline
l2 &\textbf{0.116}&    0.235&   0.691 \\
l2+gan &0.224 &    0.275 &    0.713 \\
l2+gan+df  &0.131 & \textbf{0.212} & \textbf{0.619}\\ \hline
\end{tabular}
\vspace{1mm}
\caption{Effects of different losses.} \label{fig:losses}
\vspace{-3mm}
\end{table}


\section{Conclusion and Future Work}
This paper presents `` \OURS,'' an end-to-end framework for estimating high dynamic range illumination maps for a selected pixel in a low dynamic range image of an indoor scene.  We propose to decompose the task into subtasks and train a network module for: 1) inferring 3D scene geometry, 2) warping observations to illumination maps, 3) estimating unobserved illumination, and 4) mapping LDR to HDR.  Experiments show that we can train a network with this decomposition that predicts illumination maps with better details and accuracy than alternative methods.
While ``\OURS'' is able to improve the accuracy of existing methods, it is still far from perfect. In particular, it often produces plausible illumination maps rather than accurate ones when no lights are observed directly in the input.
Possible directions for future work include explicit modeling of surface material and reflective properties and exploring alternative 3D geometric representations that facilitate out-of-view illumination estimation through whole scene understanding.

\mypara{Acknowledgments}
We thank Paul Debevec, John Flynn, Chloe LeGendre, and Wan-Chun Alex Ma for their valuable advice, Marc-André Gardner for producing results for comparison, Matterport for their dataset, and NSF 1539014/1539099 for funding.

{

\small
\bibliographystyle{ieee}
\bibliography{main}
}

\newpage

\appendix
\section{Appendix}
In this supplementary material, we provide more details about our algorithm, evaluation, and additional results. 

\section{Network Architecture Details}
We implement our network architecture in Pytorch. 
The input image's resolution is 256$\times$320 pixels, and the output illumination map's resolution is 160$\times$320 pixels. Following sections provide additional details about each sub-module's network architecture.

\mypara{Geometry estimation module} 
The geometry estimation module takes a single RGB image as input and outputs a dense pixel-wise prediction of the visible 3D geometry.
This module uses the U-Net structure \cite{ronneberger2015u} with two stream output, one for surface normal and one for plane distance. 
Let C(c,k) denotes Convolution-ReLU layer with  k filters with c channels, DC(c,k) denotes a Downsample-Convolution-ReLU with a downsample stride equals to 2, and UC denote a Upsample-Convolution-ReLU layer with a upsample stride equals to 2, the whole network module is defined as follow:

C(3,64) - DC(64,128) - DC(128,256) - DC(256,512) - DC(512,1024) - UC(1204,512) - UC(512,256) - UC(256,128) - UC(128,64) - UC(64,64)- [C(64,64) - C(64,64) - C(64,64) - C(64,64) - C(64,x)]x2

The number of channels ($x$) of the last layer depends on the stream's mortality:  for normal stream $x=3$, for plan distance stream $x=1$.

The PN-Layer then takes in a predicted normal map and plane distance map and calculates the final 3D point location for each pixel. If the pixel normal prediction is $\vec{n} =  (n_x,n_y,n_z)$ in camera coordinate (normalized to be unit length), the plane distance prediction is $p$, the camera intrinsics matrix is $K = [f_x,0,c_x;0,f_y,c_y;0,0,1] $,  and the 2D pixel location is $(x_i,y_i, 1)$, assuming the camera is at the origin, then the computed 3D point location $\vec{P} = (x,y,z)$ is 
$\vec{P} =-\frac{p}{\vec{v} \cdot  \vec{n} } \vec{v}$, where 
$\vec{v} = (\frac{x_i-c_x}{f_x},\frac{y_i-c_y}{f_y}, 1)$

\mypara{LDR completion module} 
takes the warped observation pixels as input and outputs a dense pixel-wise prediction of illumination for the full panoramic image including both observed and unobserved pixels. 
This module consists of a generative network and a discriminative network. 
The generative network is a fully convolutional ResNet50 model \cite{he2016deep}, with all the convolution layers replaced with distortion-aware convolutional filters \cite{tateno2018distortion}.
The discriminator's network architecture is defined as follow: C(3,64)- C(64,128) -  C(128,256) - C(256,512).

\mypara{LDR-HDR Estimation module} 
takes the predicted LDR illumination as input and outputs a dense pixel-wise prediction of HDR illumination intensities representing incident radiance in every direction at the target locale. 
This module uses the U-Net structure \cite{ronneberger2015u} that is defined as follows: 

C(3,64) - DC(64,128) - DC(128,256) - DC(256,512) - DC(512,1024)  - UC(1204,512) - UC(512,256) - UC(256,128) - UC(128,64) - UC(64,64)- C(64,64) - C(64,64) - C(64,64) - C(64,64) - C(64,x).

\section{Training details.}
For all modules, we randomly initialize all layers by drawing weights from a Gaussian distribution with mean 0 and standard deviation 1.  
We first pretrain each network module individually for five epochs with batch size equals to 4 with learning rate equals to 0.001. We then fine-tune the whole network end-to-end for 10,000 iterations with batch size equals to 1 and learning rate equals to 0.0001. 
The following sections describe the details of these pre-training and fine-tuning processes. 

\mypara{Geometry estimation module} 
For the geometry estimation module, during both pre-training and fine-tuning,  we directly use the color image as input and the surface normal, plane distance as supervision.
The surface normal, plane distance is directly computed from the depth images provided by Matterport3D dataset.

\mypara{ LDR completion module} 
The LDR completion module takes in a partial illumination map (the ``observation panorama'') as input and output a completed LDR illumination map.  During pre-training, the ``observation panorama'' is generated by warping the observed color image using the ground truth 3D scene geometry.  During Fine-tuning and testing, the input of this module is an estimated ``observation panorama'' generated by warping the observed RGB image with the 3D scene geometry output by the geometry estimation module.  

\mypara{ LDR to HDR estimation module} 
Similarly, for the LDR to HDR module, we use the ground truth LDR and HDR pairs from the dataset as training data during pre-training, and use the estimated LDR from the  LDR completion module as input during fine-tuning and testing.

\begin{figure*}[t]
\centering
    \includegraphics[width=\linewidth]{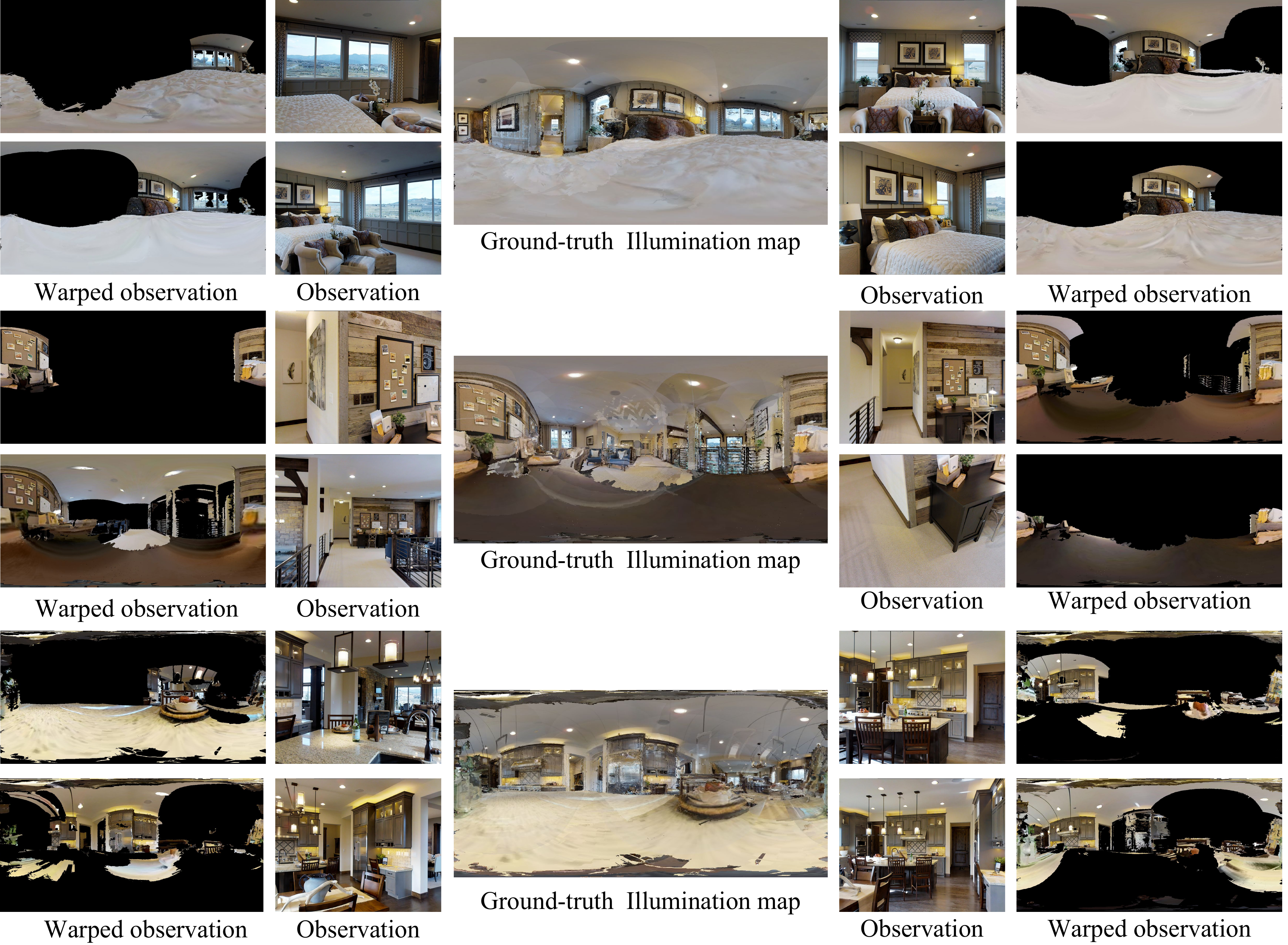}
    \caption{ \label{fig:obs} \textbf{Different observations of the same locale.} Here we show examples of different observation images and their warped partial illumination maps of the same locale. The ground truth illumination map of the locale is generated by combining these partial observations from different viewpoints together using the method described in main paper Section 4.}
\end{figure*}

\mypara{Training data}
Figure \ref{fig:data} shows two example houses in the Matterport3D dataset, and the sampled locale positions we used to generate the training data.  
Figure \ref{fig:obs} shows examples of different observation images and their warped partial illumination maps of the same target locale. The ground truth illumination map of the locale is generated by combining these warped partial observations together using the method described in main paper Section 4.

\begin{figure*}[t]
~~~~~~~~~~~~~~~~~~~~~~~floor~~~~~~~~~~~~~~~~~~~~~~~~~~~~~~~~~~~~~~~~~~~furniture~~~~~~~~~~~~~~~~~~~~~~~~~~~~~~~~~~~floor~~~~~~~~~~~~~~~~~~~~~~~~~~~~~~~~~~~~~~~furniture\\
    \includegraphics[width=0.25\linewidth]{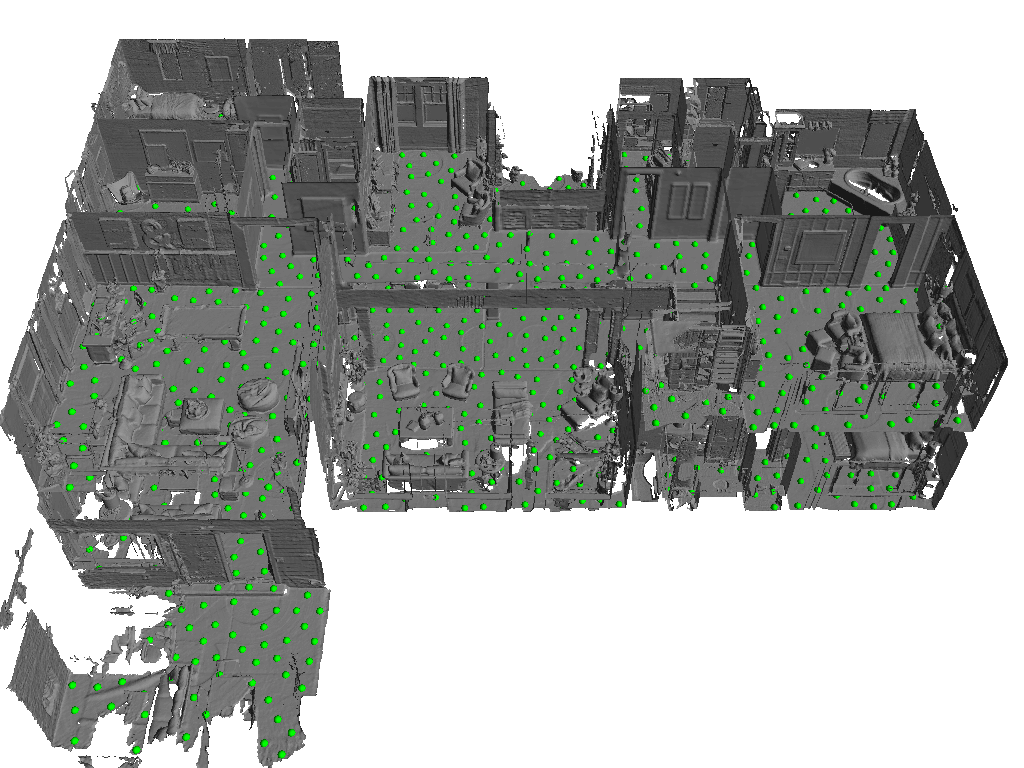}~
    \includegraphics[width=0.25\linewidth]{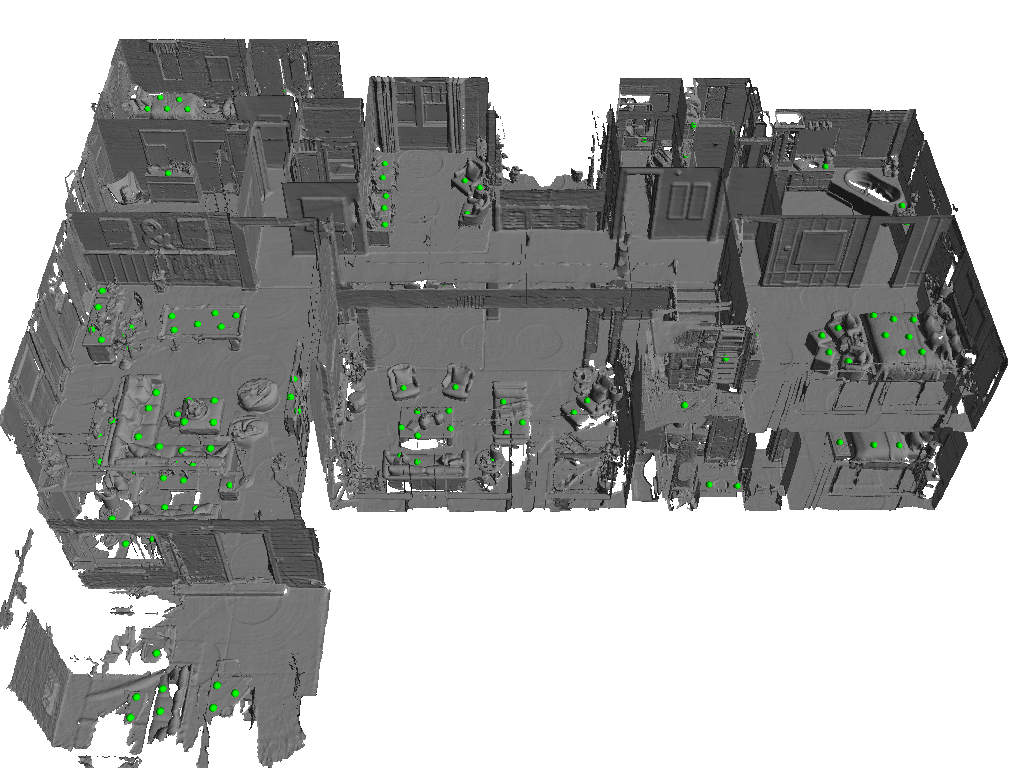}~
    \includegraphics[width=0.25\linewidth]{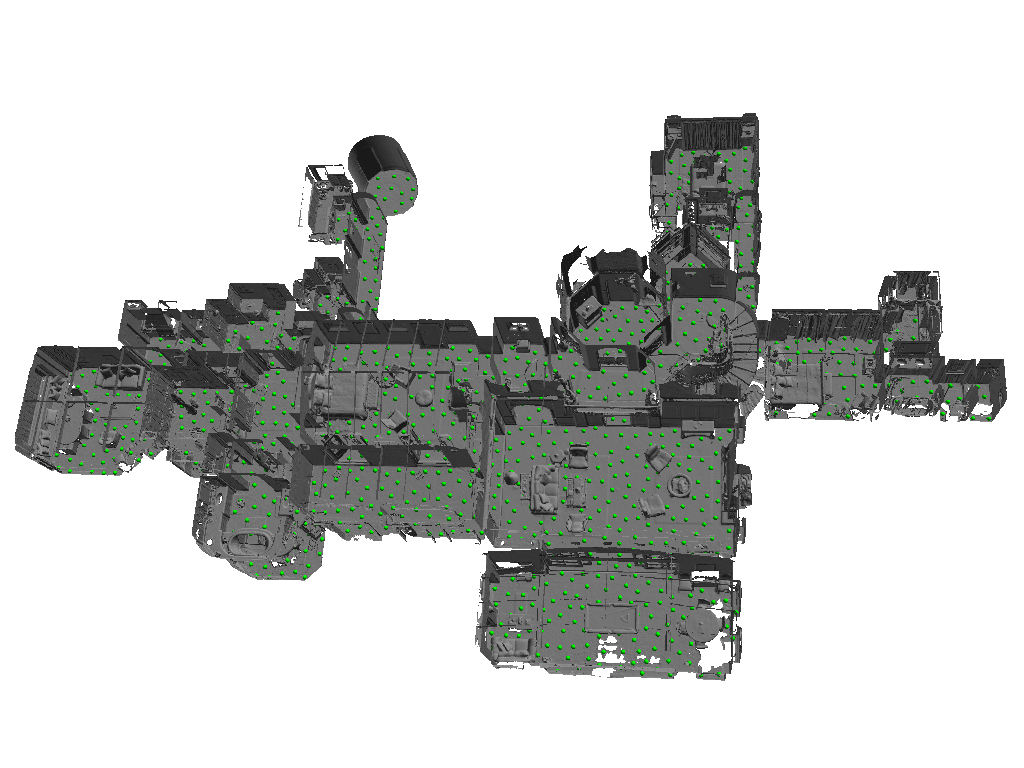}~
    \includegraphics[width=0.25\linewidth]{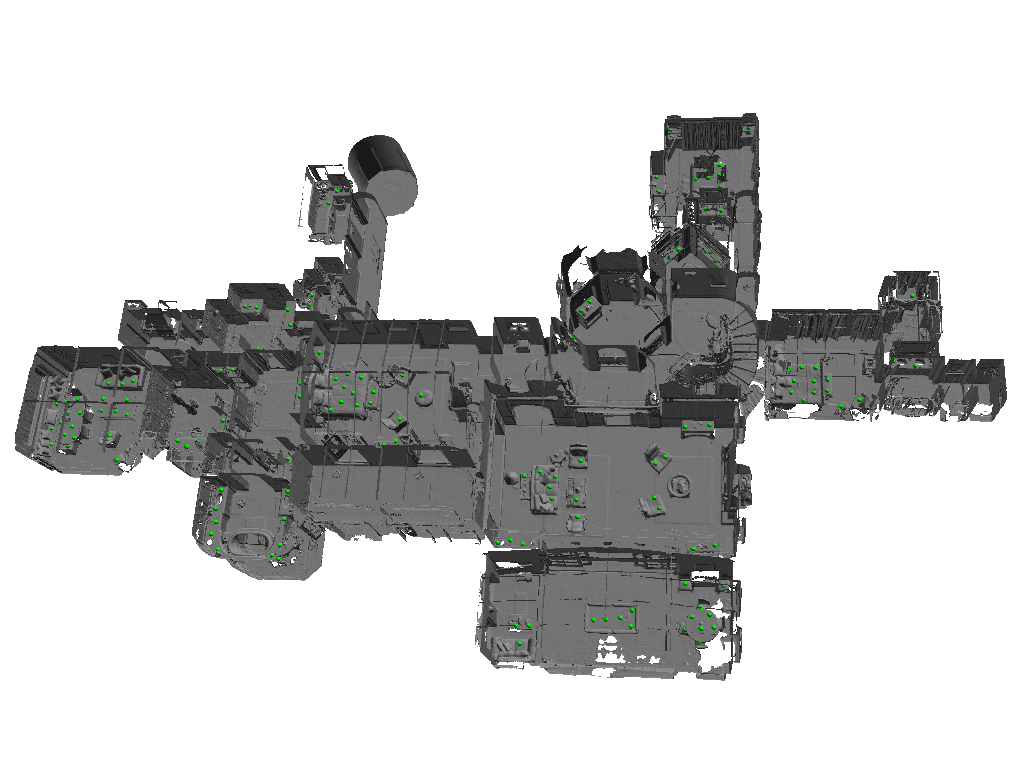}
    
    \caption{ \label{fig:data} \textbf{Sampled locale positions in Matterport3D dataset.} The sampled locale positions are visualized as green dots. The locales on the floor and furniture are visualized separately. }
    
\end{figure*}

\begin{figure*}[t]
\centering
    \includegraphics[width=0.9\linewidth]{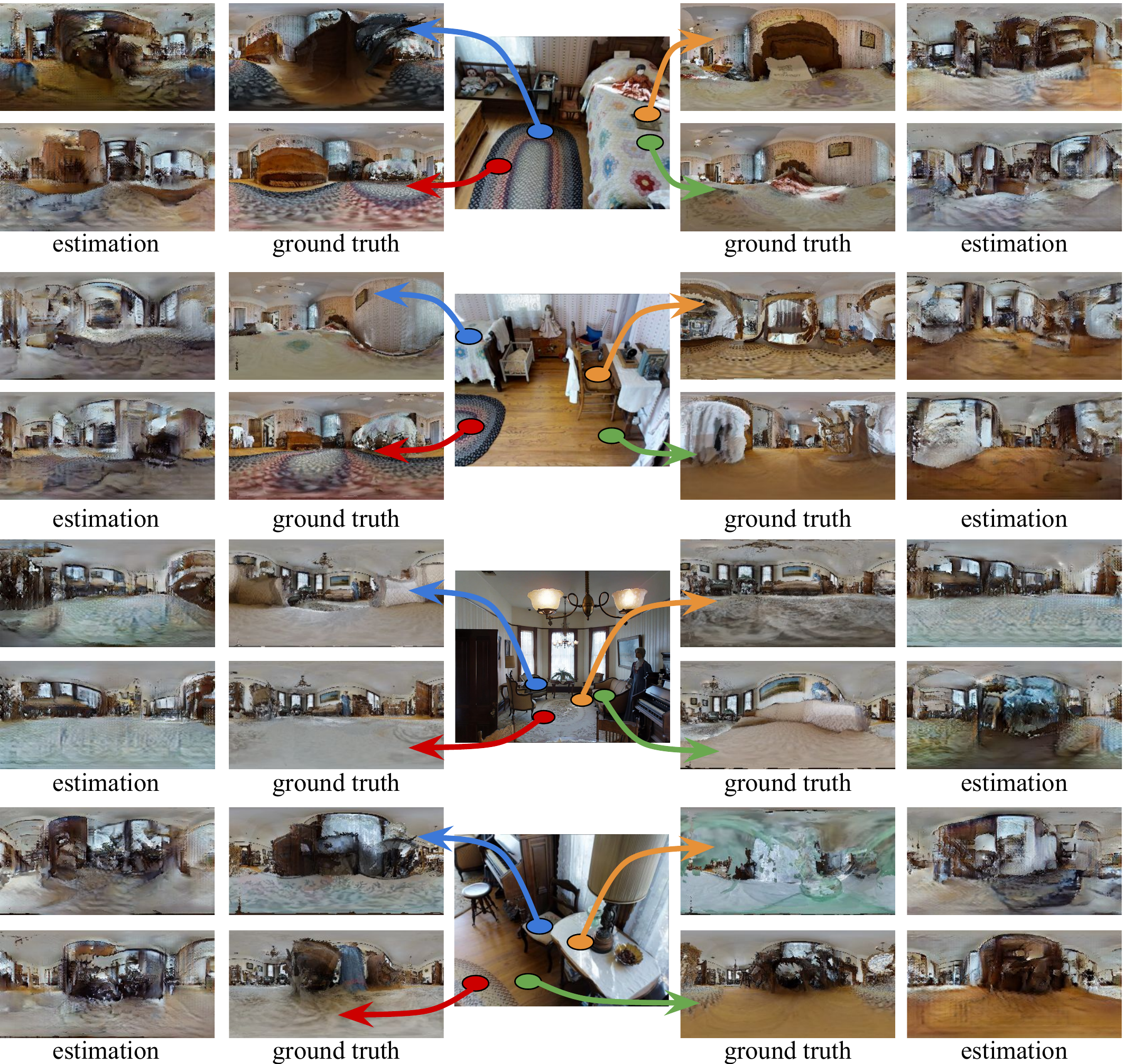}
    \caption{ \label{fig:diff_locale_more} \textbf{Spatially varying illumination.} By using the 3D geometry, we can generate ground-truth illumination for any target locale. As a result, our model is also able to infer spatially varying illumination conditioned on the location of the target pixel. 
    }
\end{figure*}

\section{Additional Results}
\begin{figure*}[t]
\vspace{-9mm}
\centering
    \includegraphics[width=0.97\linewidth]{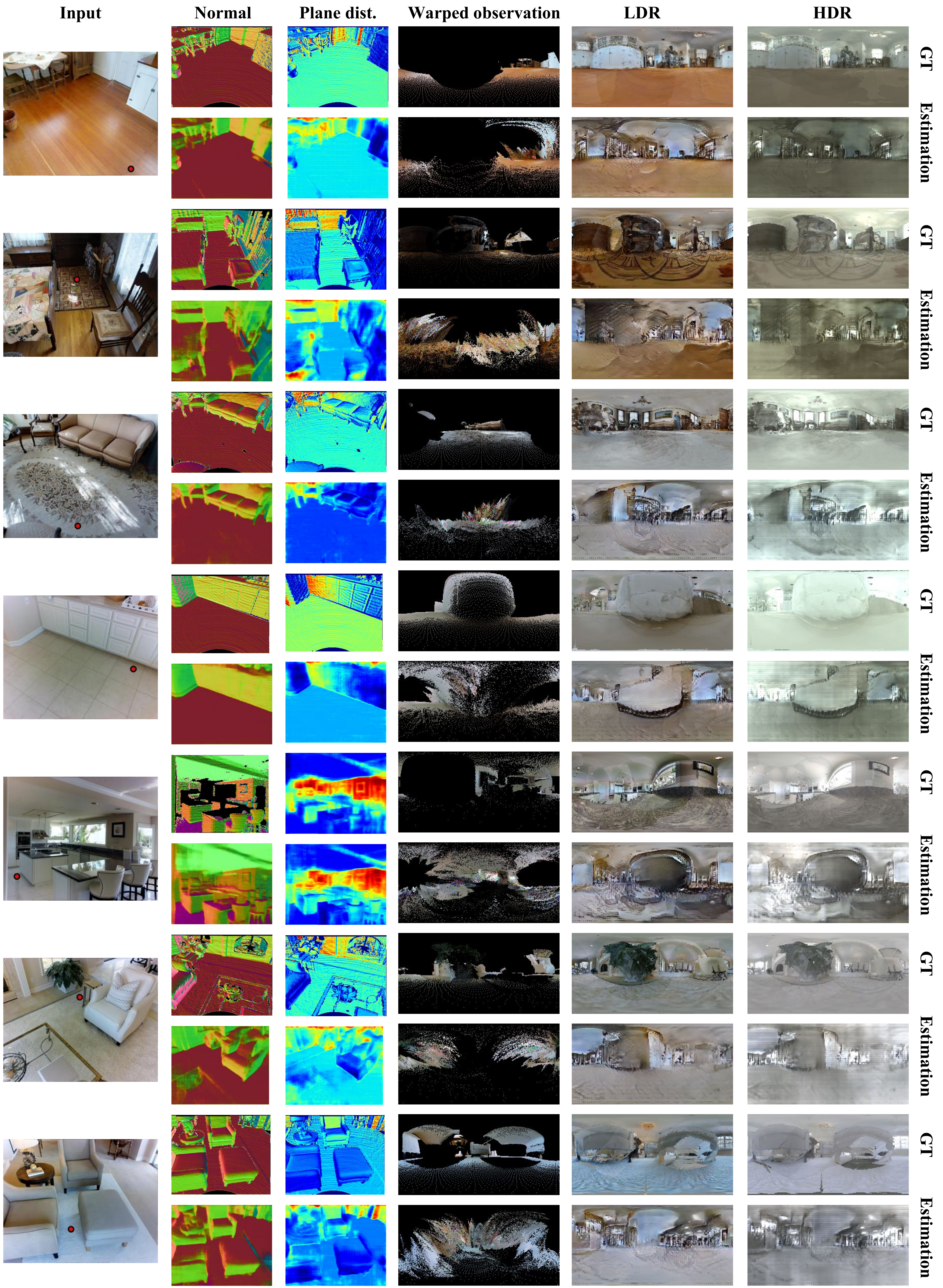}
    \caption{ \label{fig:result_more} \textbf{Qualitative results.} For each example, we show the input image and selected locale, followed by the ground truth and estimated 3D scene geometry (surface normal and plane distance), warped observation, completed LDR, and final HDR illumination. 
    }
\end{figure*}

\begin{figure*}[t]
\vspace{-9mm}
\centering
    \includegraphics[width=0.97\linewidth]{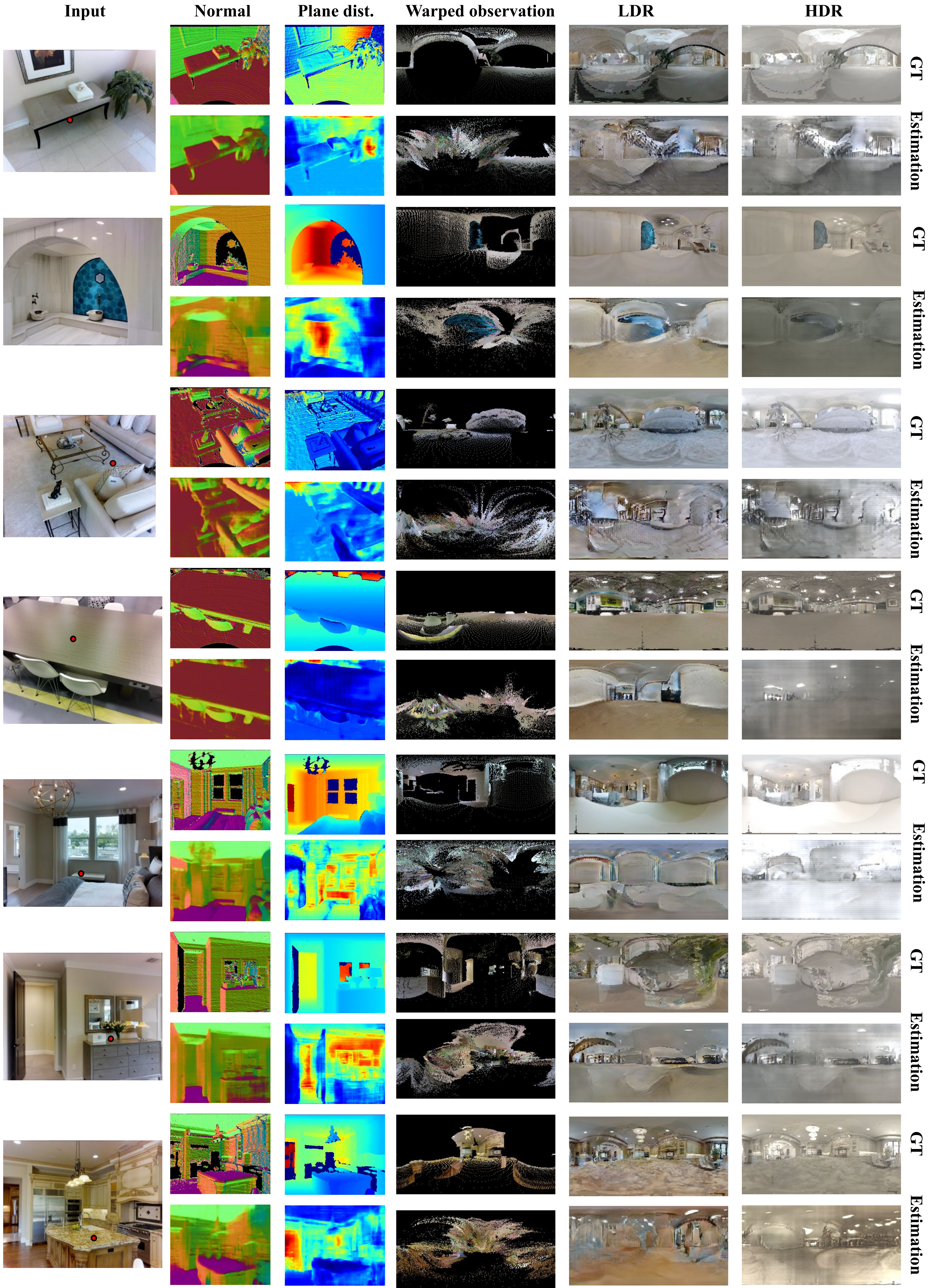}
    \caption{ \label{fig:result_more2} \textbf{Qualitative results (continued).} 
    }
\end{figure*}

\begin{figure*}[t]
\vspace{-9mm}
\centering
    \includegraphics[width=0.97\linewidth]{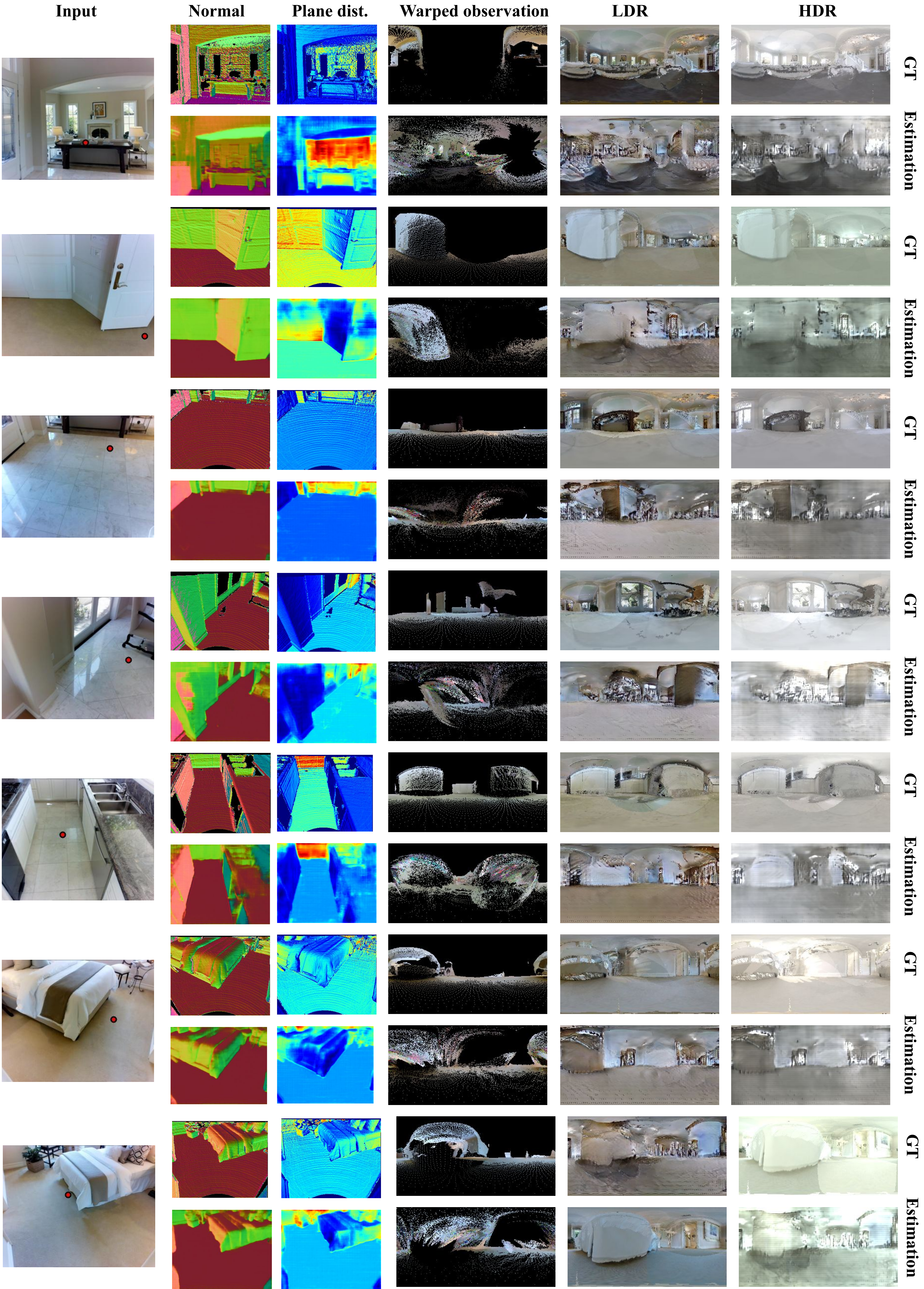}
    \caption{ \label{fig:result_more3} \textbf{Qualitative results (continued).} 
    }
\end{figure*}

Figure \ref{fig:diff_locale_more} shows more examples of spatially varying illumination. Given a single input image, we show the ground truth and estimated illumination maps conditioned on the different 2D pixels. 
Figure \ref{fig:rednering_more} shows more examples of virtual object re-lighting results using ground truth and estimated illumination maps. 
Figure \ref{fig:result_more} - \ref{fig:result_more3} shows more qualitative results. For each example, we show the input image and selected locale in Column 1. Column 2 and 3 show the ground truth and estimated 3D scene geometry visualized by the surface normal and plane distance.  Column 4 shows the warped observation using ground truth and predicted geometry. Column 5 shows the completed LDR. Column 6 shows the final HDR illumination visualized with gamma correction ($\gamma$=3.3).

\begin{figure*}[t]
\centering
    \includegraphics[width=\linewidth]{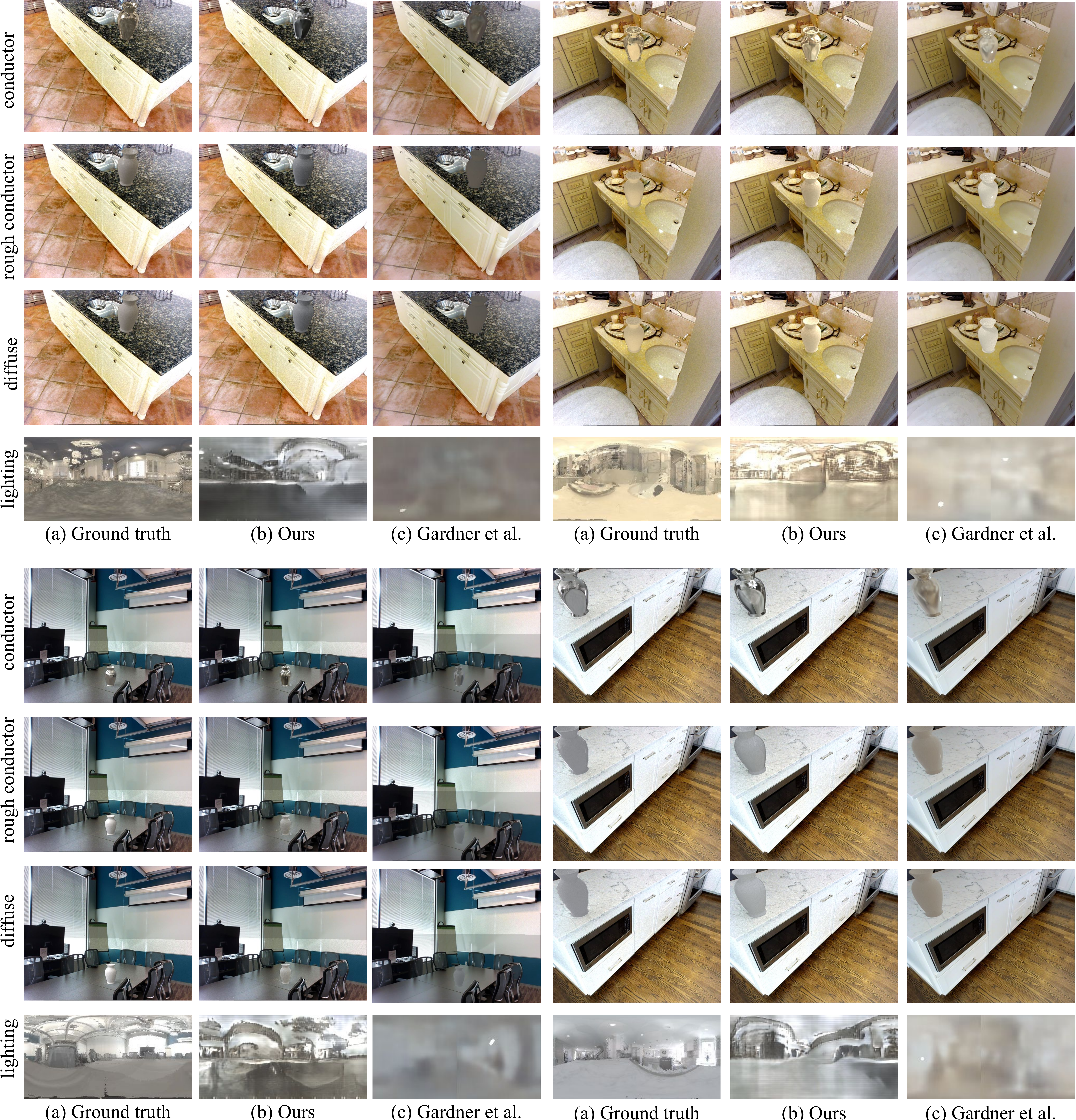}
    \caption{ \label{fig:rednering_more} {\bf Object relighting example.} Here we show more qualitative comparisons of relighting results rendered by Mitsuba using the illumination maps from (a) ground truth, (b) our algorithm, and (c) Gardner \etal.  We show images rendered with three different surface materials composited over the original observations  and the illumination maps. 
    }
\end{figure*}
\begin{figure*}[t]
    \includegraphics[width=\linewidth]{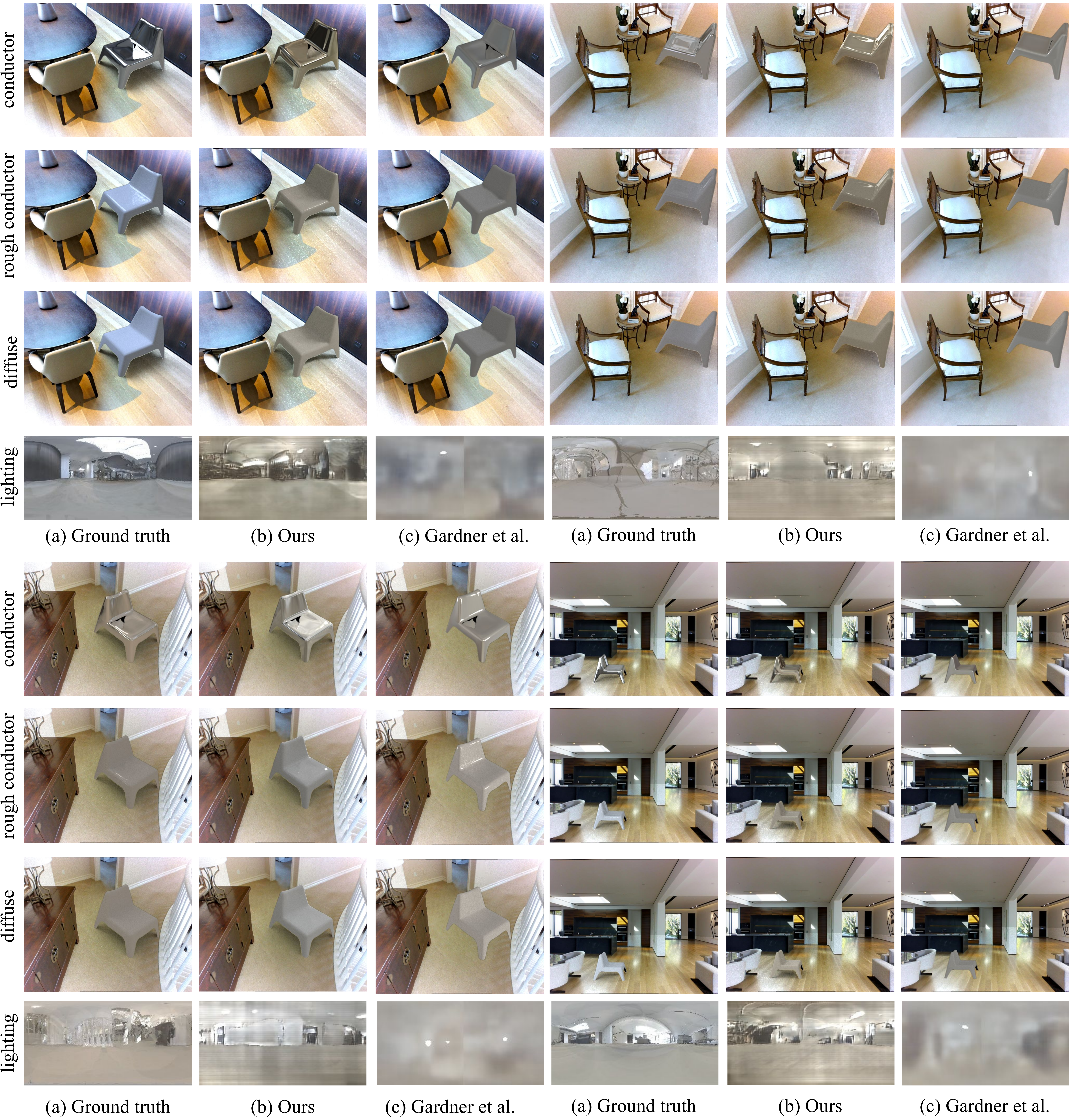}
    \caption{ \label{fig:rednering_more} {\bf Object relighting example.} Here we show more qualitative comparisons of relighting results rendered by Mitsuba using the illumination maps from (a) ground truth, (b) our algorithm, and (c) Gardner \etal.  We show images rendered with three different surface materials composited over the original observations  and the illumination maps. 
    }
\end{figure*}

\end{document}